\documentclass[lettersize,journal]{IEEEtran}
\usepackage[caption=false,font=normalsize,labelfont=sf,textfont=sf]{subfig}
\usepackage{textcomp,enumitem,amsmath,amssymb,amsfonts}
\usepackage{algorithm,algorithmic}
\usepackage{tabularray,rotating,adjustbox,multirow,array,float,graphicx}
\usepackage{xcolor,cite,etoolbox}
\UseTblrLibrary{diagbox}
\begin{document}
\title{Cyber Physical Awareness via Intent-Driven Threat Assessment:\\Enhanced Space Networks with Intershell Links}
\author{Selen~Gecgel~Cetin,~\IEEEmembership{Graduate~Student~Member,~IEEE,}~Tolga~Ovatman, \IEEEmembership{Senior~Member,~IEEE,}~and~Gunes~Karabulut~Kurt,~\IEEEmembership{Senior~Member,~IEEE}
\thanks{Received 19 June 2025; accepted 22 July 2025. The associate editor coordinating the review of this article and approving it for publication was Y. Fu. (Corresponding author: Selen Gecgel Cetin.)}
\thanks{Selen~Gecgel~Cetin is with the Department of Electronics and Communication Engineering, Istanbul Technical University, Istanbul, Turkiye and also with the Department of Electrical Engineering, Poly-Grames Research Center, Polytechnique Montreal, Montreal, QC, Canada (e-mail: gecgel16@itu.edu.tr).}
\thanks{Tolga~Ovatman is with the Department of Computer Engineering, Istanbul Technical University, Istanbul, Turkiye (e-mail: ovatman@itu.edu.tr).}
\thanks{Gunes~Karabulut~Kurt is with the Department of Electrical Engineering, Poly-Grames Research Center, Polytechnique Montreal, Montreal, QC, Canada (e-mail: gunes.kurt@polymtl.ca).}}
\maketitle
\begin{abstract}
This letter addresses essential aspects of threat assessment by proposing intent-driven threat models that incorporate both capabilities and intents. We propose a holistic framework for cyber physical awareness (CPA) in space networks, pointing out that analyzing reliability and security separately can lead to overfitting on system-specific criteria. We structure our proposed framework in three main steps. First, we suggest an algorithm that extracts characteristic properties of the received signal to facilitate an intuitive understanding of potential threats. Second, we develop a multitask learning architecture where one task evaluates reliability-related capabilities while the other deciphers the underlying intentions of the signal. Finally, we propose an adaptable threat assessment that aligns with varying security and reliability requirements. The proposed framework enhances the robustness of threat detection and assessment, outperforming conventional sequential methods, and enables space networks with emerging intershell links to effectively address complex threat scenarios.
\end{abstract}
\begin{IEEEkeywords}
Cyber physical awareness, intent-driven, intershell links, space domain awareness, threat assessment.\vspace{-2mm}
\end{IEEEkeywords}
\section{Introduction}
\begin{figure}[!t]
\center
\includegraphics[width=0.9\linewidth]{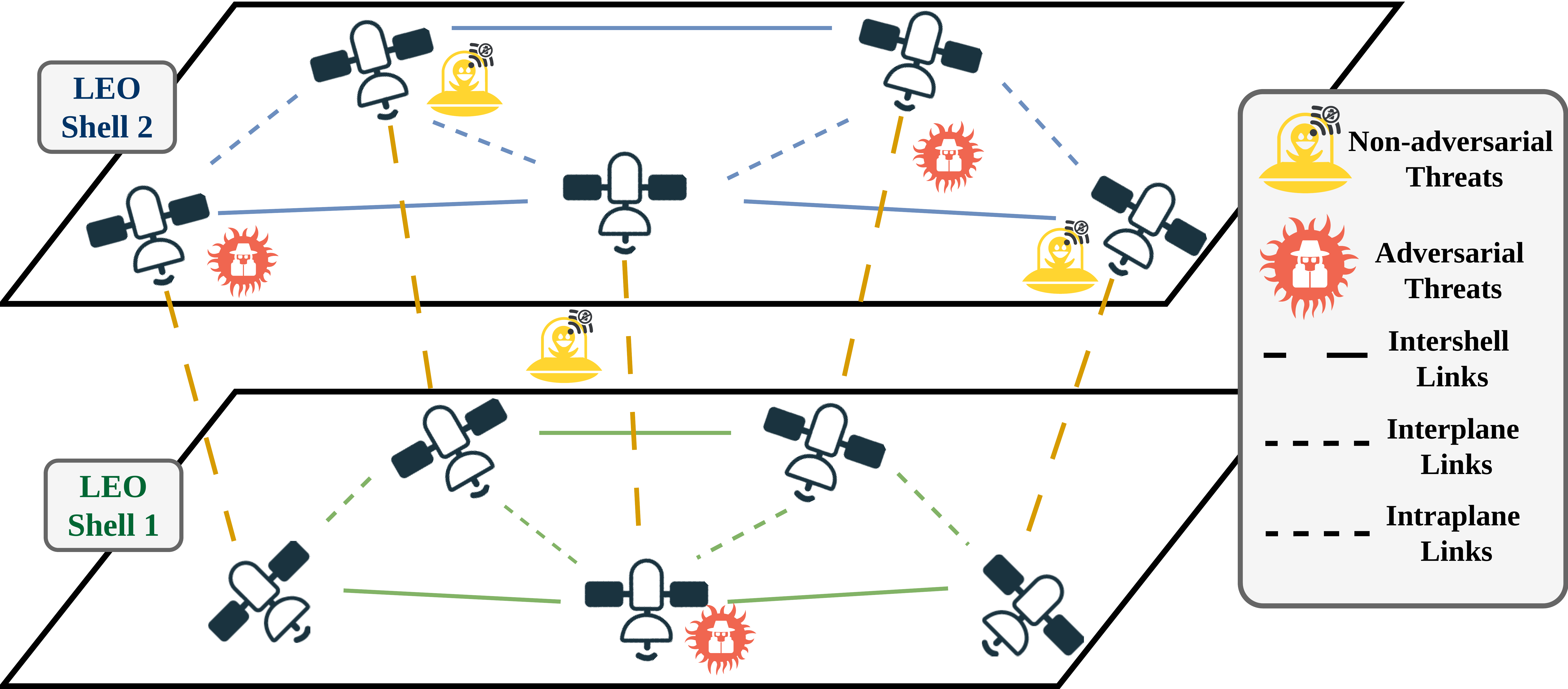}\vspace{-1.5mm}
\caption{The illustration of the use case.}\vspace{-2.3mm}
\label{fig1} 
\end{figure}

\IEEEPARstart{R}{apid} advancements in space networks over the past decade have made them a key element of several technologies. Space networks continue to expand with vast constellations of satellites positioned in various orbits, shells, and planes \cite{REF1-1}. Currently, they offer global coverage through interplane and intraplane satellite links, as well as terrestrial infrastructures. The rising demands from next-generation systems—including critical infrastructure applications like enhancing cyber-physical power grids \cite{REF-extra1}—have urged the enhancement of intershell links to optimize and build competent space networks \cite{REF1-2}. However, increased connectivity and complexity in space expose these networks to several multifaceted threats at both cyber\footnote[1]{The cyber level is the point at which threats act to jeopardize confidentiality, integrity, or availability by affecting the data or system logically, whereas the physical level is where threats impact the corporeal form of the communication signal or system.} and physical levels \cite{REF2}, as illustrated in Fig. \ref{fig1}. Therefore, cyber physical awareness (CPA) as part of the space domain awareness (SDA) \cite{REF3} is essential for recognizing potential threats to the security and reliability of communication systems in space. For example, without a comprehensive CPA, a low-power jamming assault designed just to investigate network defenses may go unnoticed if it fails to create major communication problems. Similarly, a hostile spoofing signal that replaces legitimate data may be perceived as a simple reliability issue rather than a deception attempt, potentially leading to disastrous consequences \cite{ccsds1}.
\begin{figure*}[!t]
\center
\includegraphics[width=\textwidth]{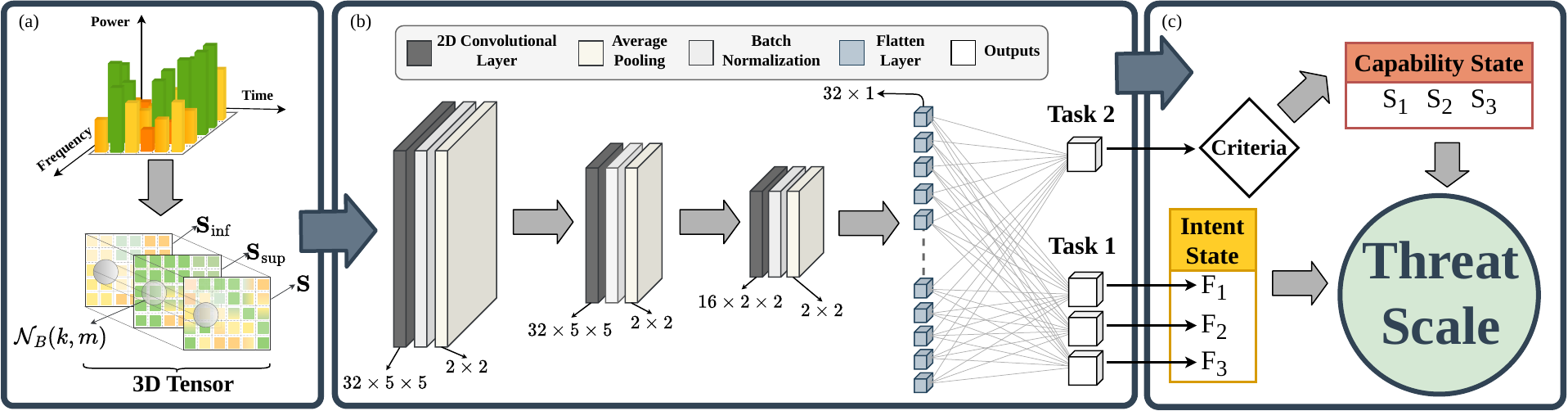}
\caption{The workflow of the proposed intent-driven CPA framework. (a) Feature representation of the CO-OFDM communication signal with Algorithm \ref{algorithm}. (b) Architecture of the designed neural network for the multitask learning. (c) Threat assessment.}\vspace{-2mm}
\label{fig2}
\end{figure*}

Security and reliability threats are generally addressed individually or in a cause-and-effect relationship. For example, security threats are frequently detected using reliability criteria, whereas reliability threats are typically detected using metrics based on signal strength or error rates \cite{REF4,REF5,REF6}. This approach performs well because the solutions and criteria are specifically arranged for the threat and its application by focusing directly on the eventual consequences rather than the underlying objectives, as seen in \cite{REF4}. Consequently, if communications are sustained without substantial errors or failures as per these criteria, the system remains completely oblivious to security threats with malicious intent. In other words, the precision of such solutions falls short when confronted with obfuscated threats due to their limited intuitive understanding.

A notable example is provided in \cite{REF5}, where a well-known learning architecture is trained to recognize four threat signals through spectrograms. Although the model achieves superior detection performance among these known threats, it is unable to distinguish unknown threats from the learned ones. In addition to these weaknesses, the solution depends exclusively on final metrics and is more vulnerable to sophisticated threats, thereby compromising CPA. Even if the presence of an unknown threat can be detected, it does not simplify the classification of its nature as either security- or reliability-relevant, since the interference may originate from significant noise, a jamming attack, or unintentional interference. The solution in \cite{REF6} employs reliability metrics to classify some security threats and detect anomalies based on these threats. The performance is effective when the relevant metrics do not overlap; however, it is limited to only known security threats. \textit{The dilemma here is: what if the threats only compromise the reliability (not security) or are unintentional?} So, dealing with security and reliability issues separately, with overspecified solutions for threats, leaves the system open to unknown risks, especially as threats become more complex and diverse.

To ensure robust CPA in space, communication systems should be multitask-oriented, which is consistent with state-of-the-art security research in other complex domains like the IoT \cite{multitask2}, but they also must assess security and reliability risks jointly \cite{multitask1}. By examining the underlying motivations of threats, we can achieve a more comprehensive understanding of their distinctions, which is crucial for developing effective strategies to mitigate risks and protect communication systems. We can address potential vulnerabilities at the cyber and physical levels by heuristically assessing threats, thereby enhancing CPA for the sustainable SDA. In this letter, we present a framework with novel methods to detect imperceptible threats, as shown in Fig. \ref{fig2}. Our contributions are listed as follows.
\begin{itemize}
\item We established a holistic CPA paradigm for enhanced space networks by addressing that threats must be assessed via their capability and intent, treating security and reliability as interdependent concepts.
\item We proposed a feature representation algorithm that creates textural signatures of signals using local supremum and infimum maps. This technique enables threat identification even when their signals are undetectable or indistinguishable via classical metrics.
\item We designed a purpose-built multitask learning architecture that assesses capability (via regression) and intent (via classification) in parallel. This structure creates a mutual dependence between the tasks, enabling a more robust and unified threat perception.
\item We introduced a custom loss function that intelligently balances both tasks' objectives. It uses focal loss to overcome class imbalance for the intent task and an uncertainty-weighting strategy to automatically tune the contribution of the capability task.
\item We developed a unified threat assessment that jointly processes both task-outputs. In contrast to conventional sequential baseline methods, this synergistic approach remedies the critical vulnerability in detecting stealthy and low-capability threats.
\end{itemize}
The feature representation algorithm effectively captures the key characteristics of the data. The neural network exhibits high performance, with $98.74\%$ accuracy for capability estimation and $99.89\%$ for intent classification. Following these stages, the proposed CPA achieves high performance for threat assessment with an overall accuracy of $98.62\%$ by overcoming the limitations of conventional cascaded approaches. This demonstrates the framework's potential to detect and grade threats across threat scales for enhanced space network operations.
\section{Problem Statement and Intent-Driven\\Threat Models}\label{Section3}
\subsection{Problem Statement}
The conventional approach to threats in communication systems, whether at the cyber or physical level, is to characterize them under security and reliability concerns individually. Jamming, spoofing, and eavesdropping are all treated as security threats, whereas system instabilities that cause errors or outages are viewed as reliability threats. As a result, strategies and techniques in the literature have developed criteria specific to each threat type and scenario, even though they employ common metrics such as bit error rate (BER), received signal strength indicator, signal-to-noise ratio, and outage incidence. Because security and reliability are interdependent, this isolated evaluation leads to two main problems that become clear when considering practical examples. For instance, a jamming assault aiming to disrupt communication will likely affect both security and reliability metrics. Although it impacts both aspects, if we classify it solely as a security threat, we risk overlooking the reliability issues. The successful detection might ensure the mitigation of potential risks; however, a failed or partially effective attack may remain undetected if the criteria focus only on threat capabilities.

Similarly, a reliability threat, such as a noise source surrounding the physical system, could be mistakenly identified as a jamming attack. These false positives and negatives in detection arise from current strategies that mainly examine threat capabilities via common metrics for security or reliability individually and do not account for the threat’s underlying intentions. Another example is a spoofing attack, which attempts to deceive the legitimate receiver. If we investigate this attack solely based on the final results—for example, a deceptive signal causing the receiver to decode incorrectly leads to a high BER—we may misinterpret its true impact and assess it merely as a disruptive event rather than as a targeted deception. This misinterpretation arises as current assessments focus on capabilities without probing the intended objectives.\vspace{-2mm}
\subsection{System Model}
We consider that legitimate satellites are positioned at different Low Earth Orbit shells and communicate via optical link, as shown in Fig. \ref{fig1}. They employ coherent optical orthogonal frequency division multiplexing (CO-OFDM) to achieve high spectral efficiency and to ensure robust sensitivity \cite{fso_book1}. The communication signal is conveyed through the CO-OFDM transmitter, optical channel, and CO-OFDM receiver. 
\subsubsection{Channel model}
The line-of-sight does not cross the lower atmosphere since satellites are high enough. Therefore, the optical link only includes pointing and the beam divergence losses. The composite channel gain is formulated at discrete-time as
\begin{equation}\label{hgain}
 h(n) = \sqrt{G_t G_r \eta_t \eta_r L_\mathrm{path}(n) L_\mathrm{point}},
\end{equation}
where $\eta_t$ and $\eta_r$ denote the transmit and receive optics efficiency, respectively. The free space path loss is $L_\mathrm{path} (n)= \left ( \lambda / 4\pi d(n) \right) ^{2}$ where $\lambda$ denotes the wavelength and $d(n)$ is the distance at time $n$. The transmitter gain $G_t$ and the receiver gain $G_r$ are calculated with $G = \left ( \pi D / \lambda \right) ^{2}$, where $D$ is the aperture diameter. The pointing loss due to jitter is given as $L_\mathrm{point} = \exp \left(-8\theta_\mathrm{jitter}^2 / \theta_\mathrm{div}^2 \right)$, where $\theta_\mathrm{jitter}$ and $\theta_\mathrm{div}$ are beam jitter and transmitter beam divergence angles in radians \cite{fso_book1}.
\subsubsection{Signal model}
The quadrature amplitude modulated symbols are mapped to the subcarriers of a CO-OFDM symbol as $ {X} =\bigl[ X(0), \ X(1),\ \cdots, \ X(N-1)\bigl]$, where $N$ is the number of subcarriers. The base-band equivalent signal in the time domain is obtained with the inverse discrete Fourier transform as
\begin{equation}
x(n) = \frac{1}{\sqrt{N}} \sum_{k=0}^{N-1} X(k) e^{j2\pi n k/N}.
\end{equation}
Then, the cyclic prefix (CP) with the length of $N_{CP}$ is appended to prevent inter-symbol interference.
\subsection{Intent-Driven Threat Models}
By handling security and reliability in isolation and focusing on only threat capabilities via specified criteria, current solutions may overlook or misidentify the genuine objectives and overall impact of the threats. A true threat assessment must investigate both the capability and the intentions of the threat, addressing security and reliability together to achieve a comprehensive evaluation. Therefore, we considered potential vulnerabilities in optical intershell links \cite{optical_threats} and modeled them based on their true intentions for legitimate transmission.
\subsubsection{Non-Adversarial Threat}
This model considers that there are no adversarial threats; hence, security is not a problem at either the cyber or physical levels. However, unintentional threats to reliability may interfere with the legitimate signal or cause errors due to system imperfections. We present the received signal below, considering only channel and thermal noise, in the absence of any intervention.
\begin{equation}
\label{eq_no}
y(n) \;=\; h_{\ell}(n)\,x(n) \;+\; w(n),
\end{equation}
where $h_{\ell}(n)$ is the channel gain of the legitimate link. $w(n)\sim\mathcal{CN}\!\bigl(0,\sigma_{w}^{2}\bigr)$ is the additive white Gaussian noise that originates from the amplified spontaneous emission and thermal noise. Even in the absence of threats, the communication signal can fluctuate significantly in time due to mobility, resulting in inconsistent reliability measures.
\subsubsection{Adversarially Disruptive Threat} This model states that the threat signal intentionally interferes with legitimate signals and disrupts communication. The receiver cannot accurately decode the transmitted signal due to the disruptive signal. It threatens both security and reliability at cyber and physical levels. We model the threat like traditional jamming with an obfuscatory factor \cite{REF-extra2}, which complicates detection by standard methods relying on reliability metrics like hypothesis tests. As stated previously, these methods mostly infer the absence of threats if the specified metrics and criteria do not indicate a notable issue. Even if the threat failed to disrupt communication but adversarial intentions exist, it still is a security threat and must be detected. With the obfuscatory factor, we can investigate the detection performance of our proposed framework under the mask of a potential security breach. The received signal is defined as
\begin{equation}
y(n) \;=\; h_{\ell}(n)\,x(n) \;+\; h_{j}(n)\,x_{j}(n) \;+\; w(n),
\end{equation}
where $h_{j}(n)$ is the channel gain corresponding to the threat and $x_{j}(n)$ represents the disruptive signal as follows
\begin{equation}
x_j(n) = \alpha(n) \, j(n).
\end{equation}
Here, $j(n)\sim \mathcal{CN}\!\bigl(0,\sigma_{j}^{2}\bigr)$ and $\alpha(n) \in \left\{0,1\right\}$ represents the obfuscatory factor which is a Bernoulli distributed variable with the probability $p_{\alpha}$.
\subsubsection{Adversarially Deceptive Threat}
We model the threat that intervenes intentionally to mislead the legitimate receiver. The adversary imitates and replaces the legitimate signal with the deceptive signal. In this way, the receiver decodes malicious information $s(n)$ as if it were an authentic signal. The received signal is expressed as
\begin{equation}
 y(n) \;=\; h_{\ell}(n)\,x(n) \;+\; x_{s}(n) \;+\; w(n).
\end{equation}
The deceptive signal, like that of a spoofer \cite{REF-extra3}, is given below
\begin{equation}
 x_{s}(n) \;=\; h_{s}(n)\,s(n) \;-\; \hat{h}_{\ell}(n)\,x(n),
\end{equation}
where $h_{s}(n)$ is the channel gain of the adversary. $\hat{h}_{\ell}(n)$ denotes the estimated channel gain of the legitimate link and subject to estimation error $\xi$ as follows
\begin{equation}
\hat{h}_{\ell}(n) \;=\; h_{\ell}(n)\,\bigl(1 - \xi\bigr).
\end{equation}
This manipulation can lead to significant consequences for both reliability and security, as the receiver may act on the false information, potentially resulting in harmful decisions or actions at the cyber and physical levels.
\begin{figure}[!ht]
\centering
\includegraphics[width=\linewidth]{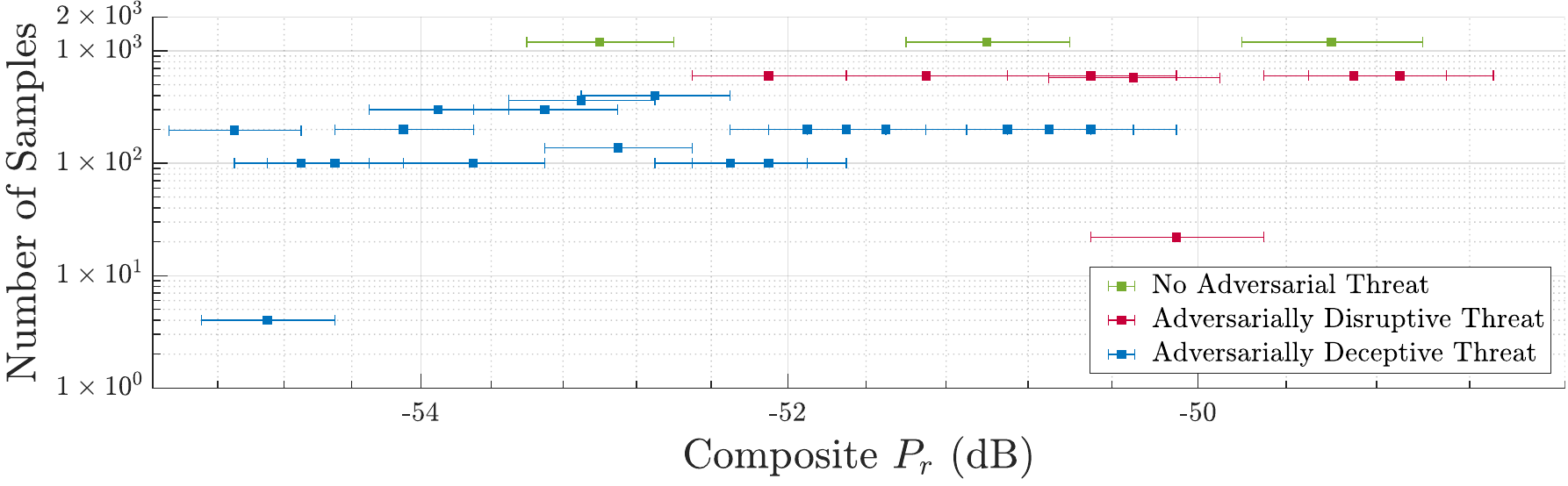}
\caption{The distribution of sub-datasets depending on the composite $P_r$.}
\label{fig3}
\end{figure}
\section{Cyber Physical Awareness Framework}
Our proposed CPA framework, shown in Fig. \ref{fig2}, approaches the threat from two perspectives: intent and capability. It proceeds methodically from the problem (use case) to the extraction of robust features (feature representation). Then it employs a multitask neural network to capture both the intent and capability of the threat. Finally, it uses the results to provide a comprehensive threat assessment. The integration of each stage allows for a more nuanced and adaptable detection against sophisticated space communication threats.
\subsection{Data Generation and Feature Representation}
Communication signals are generated according to the intent-driven threat models with the parameter settings in Table \ref{params}. The previously discussed difficulties were taken into account when determining parameter values. $p_\alpha$ and $\xi$ are set to $0.5$ and $0.3$, respectively. We compose training and test datasets with 10800 samples, and each threat model includes 3600 samples. Non-adversarial and disruptive threats are labeled with the legitimate signal BER, since they can be understood from their impact on legitimate signals. For deceptive threats, this strategy can mislead the neural network because it can disrupt legitimate signals even if its malicious information is not effectively decoded. Thus, we use the BER of the deceptive adversary for labeling, which quantifies the adversary's capability from their perspective.
\begin{table}[!b]
\begin{minipage}[b]{0.498\linewidth}
\caption{Data generation parameters.}
\resizebox{\linewidth}{!}{
\begin{tblr}{
cells = {c},
cell{1}{1} = {c=2}{}, cell{1}{4} = {c=2}{}, cell{2}{1} = {r=9}{}, cell{2}{2} = {r=2}{},
cell{2}{3} = {r=2}{}, cell{4}{2} = {r=2}{}, cell{4}{3} = {r=2}{}, cell{6}{4} = {c=2}{},
cell{7}{4} = {c=2}{}, cell{8}{4} = {c=2}{}, cell{9}{4} = {c=2}{}, cell{10}{4} = {c=2}{},
cell{11}{1} = {r=4}{}, cell{11}{4} = {c=2}{}, cell{12}{4} = {c=2}{}, cell{13}{4} = {c=2}{},
cell{14}{4} = {c=2}{},
vlines,
hline{1-2,11,15} = {-}{},
hline{3,5} = {4-5}{},
hline{4,6-10,12-14} = {2-5}{},
 }
 \text{\large Parameter}
 &         & \text{\large Unit} & \text{\large Value}  &     \\
 \begin{sideways}\text{\large Transmitter}\end{sideways}
 & $P$        & W    & Legitimate   & 0.5    \\
 &         &      & Adversarial  & 0.25, 0.5    \\
 & $d$        & km   & Legitimate   & 500, 750, 1500  \\
 &         &      & Adversarial  & 750, 1500, 3000 \\
 & $D$        & m    & 0.1    &     \\
 & $\lambda$     & nm   & 1500   &     \\
 & $N$        & -    & 512    &     \\
 & $N_\mathrm{CP}$     & -    & 64     &     \\
 & $M$        & -    & 600    &     \\
 \begin{sideways}\text{\large Receiver}\end{sideways} 
 & $\sigma_w^2$     & dB   & -56, -57  &     \\
 & $\theta_\mathrm{jitter}$  & rad     & 0.002     &     \\
 & $\theta_\mathrm{div}$  & rad     & 0.02   &     \\
 & $D$        & m    & 0.2    &     
 \end{tblr}
}
\label{params}
\end{minipage}
\hfill
\begin{minipage}[b]{0.485\linewidth}
\caption{Threat scales.}
\resizebox{\linewidth}{!}{%
\begin{tabular}{|c|c|c|c|c|} \hline
\multicolumn{2}{|c|}{\multirow{2}{*}{\begin{tabular}[c]{@{}c@{}}\text{\large Threat}\\\text{\large Scale}\end{tabular}}} 
& \multicolumn{3}{c|}{\begin{tabular}[c]{@{}c@{}}\text{\large Capability State:}\\\text{\large$S = [S_1S_2S_3]$}\end{tabular}}\\ \cline{3-5}
\multicolumn{2}{|c|}{}
&\begin{tabular}[c]{@{}c@{}}\large High\\$\mathbf{S} = [1\,0\,0]$\end{tabular}
& \begin{tabular}[c]{@{}c@{}}\large Moderate\\$\mathbf{S} = [0\,1\,0]$\end{tabular}
& \begin{tabular}[c]{@{}c@{}}\large Low\\$\mathbf{S} = [0\,0\,1]$\end{tabular}\\ \hline
\multirow{3}{*}{\rotatebox[origin=l]{90}{\mbox{\text{\large Intent State:~$\mathbf{F} = [F_1\,F_2\,F_3]$\hspace{-11.5mm}}}}}
&\begin{sideways}\begin{tabular}[c]{@{}c@{}}\hspace{-4mm}Non-Adversarial~~\\\hspace{-5mm}$\mathbf{F} = [0\,0\,1]$~~\end{tabular}\end{sideways}
&\begin{tabular}[c]{@{}c@{}}\text{\Large2}\vspace{7mm}\end{tabular}
&\begin{tabular}[c]{@{}c@{}}\text{\Large1}\vspace{7mm}\end{tabular}
&\begin{tabular}[c]{@{}c@{}}\text{\Large0}\vspace{7mm}\end{tabular}\\ \cline{2-5}
&\begin{sideways}\begin{tabular}[c]{@{}c@{}}\hspace{-2.7mm}Disruptive~~~\\\hspace{-3.2mm}$\mathbf{F} = [0\,1\,0]$~~~\end{tabular}\end{sideways}
&\begin{tabular}[c]{@{}c@{}}\text{\Large4}\vspace{5mm}\end{tabular}
&\begin{tabular}[c]{@{}c@{}}\text{\Large3}\vspace{5mm}\end{tabular}
&\begin{tabular}[c]{@{}c@{}}\text{\Large3}\vspace{5mm}\end{tabular}\\ \cline{2-5}
&\begin{sideways}\begin{tabular}[c]{@{}c@{}}\hspace{-3.6mm}Deceptive~~\\\hspace{-3.1mm}$\mathbf{F} = [1\,0\,0]$~~~\end{tabular}\end{sideways}
&\begin{tabular}[c]{@{}c@{}}\text{\Large5}\vspace{5mm}\end{tabular}
&\begin{tabular}[c]{@{}c@{}}\text{\Large6}\vspace{5mm}\end{tabular}
&\begin{tabular}[c]{@{}c@{}}\text{\Large7}\vspace{5mm}\end{tabular}\\ \hline
\end{tabular}
}
\label{states}
\end{minipage}
\end{table}

Fig. \ref{fig3} shows the composite received power ($P_r$) distribution of the test dataset that is typically between $-49$ and $-37$ dB. This phenomenon highlights the constraints in identifying threats using conventional approaches and the possibility of missing threats. To overcome these difficulties, we first propose Algorithm \ref{algorithm} for feature representation in our framework. It examines the spectrogram to extract features related to the threat capability, such as signal strengths. As spectrograms alone are insufficient to understand the threat intent, the algorithm creates new features by dilating and eroding over a defined neighborhood with radius $R_B$ (which is set to 15 in this letter). This is critical for capturing the distinct textural signatures of threats, which raw $P_r$ levels often obscure, as in Fig. \ref{fig3}. The local supremum and infimum feature maps highlight the highest and lowest intensity points, respectively. They concretize local contrast and small differences to distinguish between normal fluctuations and anomalies. They also allow the learning model to distinguish high-contrast patterns of disruptive threats like jamming from the subtle, unnatural artifacts of a deceptive signal by analyzing
\begin{algorithm}[!t]
\caption{Feature Representation}
\label{algorithm}
\begin{algorithmic}
\REQUIRE The received signal block (CP-removed) $y(n)\in\mathbb{C}^{N\times M}$.
\STATE (1) Initialize $R_B$ and the number of frequency bins in the spectrogram $K = M$.
\STATE (2) Convert $y(n)$ to a 1D vector and define the neighborhood disk:
\begin{equation*}
B = \left\{(u,v) \in \mathbb{Z}^2 : u^2 + v^2 \le R_{B}^{2}\right\}.
\end{equation*}
\FORALL{$k, m \in \mathbb{Z} $}
\STATE (3) Compute the spectrogram:
\begin{equation*}
S(k, m) = \left|\ \sum_{n=0}^{N-1} 
y(n + kN) e^{-\mathrm{j}\tfrac{2\pi}{N}mn}\right|.
\end{equation*}
\STATE (4) Compute the neighborhood:
\begin{equation*}
\mathcal{N}_B(k,m)=\left\{(i,j)\in\mathbb{Z}^2:(i-k)^2+(j-m)^2 \le R_B^2 \right\}.
\end{equation*}
\STATE (5) Compute the local supremum and infimum:
\begin{align*}
S_{\sup}(k,m) &= {\sup_{(i,j) \,\in \, \mathcal{N}_B(k,m)}} S(i,j),\\
S_{\inf}(k,m) &= \inf_{(i,j) \, \in \, \mathcal{N}_B(k,m)} S(i,j).
\end{align*}
\STATE (6) {Stack $S(k,m)$, $S_{\sup}(k,m)$, $S_{\inf}(k,m)$ into $ Z \in \mathbb{R}^{K\times M\times 3}$}.\vspace{-0.5mm}
\ENDFOR\vspace{-5.5mm}
\STATE \RETURN $Z$
\end{algorithmic}
\end{algorithm}
local signal behavior rather than its absolute magnitude. By stacking the spectrogram with these maps into a 3D tensor as shown in Fig. \ref{fig2}a, the algorithm creates a more comprehensive representation, preserving global signal characteristics and fine-grained local variations. This algorithm leverages the multitask model to mutually discern the intention and capability of threats by accenting signal characteristics within expected patterns of benign and malicious behavior.
\subsection{Multitask Learning}
The framework introduces a neural network in Fig. \ref{fig2}b within a multitask learning strategy, which is chosen to assess the dual objectives simultaneously: capability (via regression) and intent (via classification). Such an approach is critical for correctly identifying threats that are stealthy but malicious, a key vulnerability in systems that assess these risks separately. The network architecture, shown in Fig. \ref{fig2}b, comprises a shared backbone of 2D convolutional layers, which extract hierarchical patterns, followed by average pooling and batch normalization. These layers utilize ReLU activation functions and He normal kernel initializer to prevent the vanishing gradient problem. Additionally, L2 kernel regularization ($10^{-4}$) is applied to the filters to prevent overfitting. The shared-weight backbone feeds two separate output heads, creating a mutual dependence between tasks. This interdependence compels the model to learn in sync and be robust for both discerning threat intent and assessing its capability.

A custom loss function is necessary for balancing the dual-objectives of multitask learning. For the classification (Task 1), we employ categorical focal loss \cite{focal} to prevent the inherent class imbalance in threat detection by focusing the training on hard-to-classify samples, calculated as
\begin{equation}
\mathcal{L}_{\mathrm{Cl}} = \frac{1}{\Upsilon} \sum_{\upsilon=1}^\Upsilon 
\Biggl(-\sum_{c=1}^C q_{\upsilon,c} \bigl(1 - \hat{q}_{\upsilon,c}\bigr)^{\gamma} 
\log\bigl(\hat{q}_{\upsilon,c}\bigr) \Biggr),
\end{equation}
where $\Upsilon$ is the batch size, $C$ is the number of classes, $q_{\upsilon,c}$ is the label of sample $\upsilon$, and $\gamma$ is the focal parameter controlling the focus. For the regression (Task 2), we use mean square errors to calculate loss depending on the sample BER ($\upsilon_\mathrm{BER}$) as follows.
\begin{equation}
\mathcal{L}_{\mathrm{Reg}} = \frac{1}{\Upsilon} \sum_{\upsilon=1}^\Upsilon 
\bigl(\rho_\upsilon - \hat{\rho}_\upsilon\bigr)^2, \; \text{ where } \rho_\upsilon = \log(\upsilon_\mathrm{BER}).
\end{equation}
As the class imbalance of the Task 1 is managed by the focal loss function, we set the total loss by weighting with the variance of the dataset $\sigma^2_{\mathrm{Reg}}$ due to the uncertainty of the Task 2. This allows the model to automatically tune the contribution of each task. $\mathcal{L}_{\mathrm{Reg}}$ is also regularized with an amplification factor $\varpi$ to prevent overfitting due to the logarithmic labeling. This leads to the total loss:
\begin{equation}
\mathcal{L}_\mathrm{Total} = \mathcal{L}_{\mathrm{Cl}} + \bigl(1 /{\varpi \sigma_{\mathrm{Reg}}^2}\bigr) \mathcal{L}_\mathrm{Reg}.
\end{equation}
This custom loss structure ensures that both intent and capability are learned robustly, preventing the model from sacrificing sensitivity to one task for the other. The architecture is optimized using the ADAM optimizer with a learning rate of~$10^{-4}$.
\subsection{Threat Assessment}
The last stage of the CPA framework evaluates both the security and reliability consequences at the cyber and physical levels, as shown in Fig. \ref{fig2}c. Threat states are defined using the multitask learning stage's one-hot encoded outputs. The classification task directly produces the intent state of threat ($\mathbf{S}$), but the regression task's estimated BER values are categorized with system-specific criteria. These criteria are adaptable and can be tailored to requirements of communication systems. We define the criteria for the capability state as follows: high BER as larger than $10^{-2}$, low BER as less than $10^{-4}$, and values in between as moderate BER. According to these criteria, the threat capability has been assessed as high, moderate, or low. According to our labeling process, a high BER for non-adversarial and disruptive threats corresponds to a high capability level, whereas a high BER for deceptive threats implies a poor capability state. We scale the threat based on its intent and capability states as shown in Table \ref{states}.
\subsection{Benchmark Study}
To evaluate the performance of our multitask learning strategy for the CPA framework and the reasons why security and reliability threats should not be assessed separately, we establish a conventional benchmark based on a sequential threat assessment workflow. While our CPA framework with multitask learning, employing a mutually dependent learning policy for task engagement, produces one learning model, this benchmark utilizes two single-task models—a capability-only regression model and an intent-only classification model—which are trained independently. To ensure a fair comparison, all models employ the identical neural network backbone, hyperparameter settings, and datasets. Here, the capability-only model first analyzes the received signal and estimates BER. If the predicted BER exceeds a predefined threshold ($\vartheta_\text{BER}$), indicating a potential threat, only then is the intent-only model invoked to perform an analysis to classify the threat intent. While logical, this sequential approach has inherent limitations, such as an inability to detect low capability but adversarial threats. Our proposed multitask framework is designed to overcome this by assessing capability (linked to reliability) and intent (linked to security) simultaneously.
\begin{table}[!b]
\centering
\caption{Task-level results of the multitask and single-task models.}
\label{table3}
\resizebox{\linewidth}{!}{%
\begin{tblr}{
row{even} = {c},  row{3} = {c},  row{5} = {c},  row{7} = {c},  row{9} = {c},
cell{1}{1} = {c},  cell{1}{2} = {r},  cell{1}{4} = {c},  cell{1}{5} = {c},  cell{1}{6} = {c},  cell{1}{7} = {c},
cell{2}{1} = {r=4}{c},  cell{2}{2} = {r=2}{c},    cell{4}{2} = {r=2}{c},
cell{6}{1} = {r=4}{c},  cell{6}{2} = {r=2}{c},  cell{8}{2} = {r=2}{c},
vlines,  vline{1-2} = {1}{white},  hline{1} = {1-2}{white},  
hline{1} = {3-7}{},  hline{2,6,10} = {-}{},  hline{3,5,7,9} = {3-7}{},  hline{4,8} = {2-7}{},
}
&    & \diagbox{\text{Task~Targets}}{\text{Threat~Models}}  &\text{Non-Adversarial}    & \text{Disruptive}   &\text{Deceptive} & \text{Overall}             \\
\begin{sideways}\text{\large{Multitask}}\end{sideways}
& \begin{sideways}Loss\end{sideways} & Intent (Task 1)        &\text{$9.04\times10^{-4}$}&\text{$6.34\times10^{-5}$} & \text{$0.0017$}  & \text{$8.79\times10^{-4}$ } \\
&                                    & Capability~(Task 2)    & $0.0266$                 & $0.0063$            & $0.0365$        & $0.0232$            \\
& \begin{sideways}Acc.\end{sideways} & Intent~(Task 1)        & $100$                    & $100$               & $99.67$         & $99.89$             \\
&                                    & Capability~(Task 2)    & $96.53$                  & $100$               & $99.67$         & $98.74$             \\
\begin{sideways}\text{\large{Single-task}}\end{sideways} 
& \begin{sideways}Loss\end{sideways} & Intent-only            & $0.0001$                & $0.0000$             & $0.0015$        & $0.0163$            \\
&                                    & Capability-only        & $0.0144$                & $0.0042$             & $0.0482$        & $0.0225$           \\
& \begin{sideways}Acc.\end{sideways} & Intent-only            & $99.97$                 & $100$                & $99.97$         & $99.97$             \\
&                                    & Capability-only        & $98.07$                 & $100$                & $99.59$         & $98.84$             
\end{tblr}
}
\end{table}
\begin{table}[!t]
\centering
\caption{Comparative performance of multitask and sequential assessments.}
\label{table4}
\resizebox{\linewidth}{!}{%
\begin{tblr}{
  cells = {c},
  cell{1}{1} = {r=2}{c},
  cell{1}{2} = {r=2}{c},
  cell{1}{3} = {c=3}{c},
  cell{1}{6} = {c=3}{c},
  cell{1}{9} = {r=2}{c},
  cell{4}{1} = {r=3}{c},
  vlines,
  hline{1,3-4,7} = {-}{},
  hline{2} = {3-8}{},
  hline{5-6} = {2-9}{},
}
{Learning\\Policy} & $\large\vartheta_\mathrm{BER}$ & \text{Precision}       &            &           & \text{Recall}          &            &           & {Overall\\Accuracy}\\
                   &                        &\text{Non-Adversarial} & \text{Disruptive} & \text{Deceptive} & \text{Non-Adversarial} & \text{Disruptive} & \text{Deceptive} &                     \\
{Multitask}     & $-$                    & $96.53$         & $100$      & $99.67$   & $96.53$         & $100$      & $99.67$   & $98.62$             \\
{Single-\\task}    & $10^{-2}$              & $71.57$         & $100$      & $100$     & $100$           & $83.33$    & $77.17$   & $86.80$             \\
                   & $10^{-3}$              & $88.06$         & $100$      & $100$     & $100$           & $100$      & $86.88$   & $95.50$             \\
                   & $10^{-4}$              & $92.17$         & $100$      & $100$     & $100$           & $100$      & $91.58$   & $97.18$             
\end{tblr}
}
\end{table}
\section{Results}
In this section, we first show the performance of each learning model, then compare the overall performance of single-task and multitask learning strategies, and finally discuss the proposed framework results for each threat scale. The task-level results for both approach-based models are detailed in Table \ref{table3}, which demonstrates the high proficiency of baseline models when they are isolated and focused on specific tasks. The intent-only single-task model achieves an overall classification accuracy of $99.97\%$, while the capability-only model achieves a categorical accuracy of $98.84\%$ in estimating BER levels. These strong individual performances confirm that the baseline system provides a rigorous and challenging benchmark. The loss values imply the nuances of each learning policy when faced with difficult threats like deceptive threats, where our framework reduces the capability loss ($0.0365$ vs. $0.0482$ for the single-task model).

The core comparison of the frameworks, presented in Table \ref{table4}, reveals the practical advantages of our approach. The results clearly indicate that our proposed multitask framework achieves a superior and robust overall accuracy of $98.62\%$, surpassing the maximum accuracy of $97.18\%$ achieved by the single-task learning framework. The performance gap mostly stems from the independently cascaded task engagement's reliance on capability (i.e., reliability metrics) as a preliminary check. This vulnerability becomes obvious when handling deceptive and non-adversarial threats. Particularly, at a BER threshold of $10^{-2}$, the single-task framework's recall for this class plummets to a critically low $77.17\%$. This leads to a significant security vulnerability, as adversaries who do not meet the capability threshold are incorrectly assessed as non-adversarial. The cascaded system faces a trade-off due to non-adversarial threats; boosting its sensitivity to deceptive attacks by reducing the BER threshold substantially diminishes its precision on non-adversarial signals (from $92.17\%$ to $71.57\%$), hence raising the risk of false alarms. In contrast, our framework overcomes this limitation by design, maintaining high precision and recall across all threat model simultaneously, proving its effectiveness in identifying stealthy attacks regardless of their immediate impact.

\begin{figure}[!ht]
\center
\includegraphics[width=\linewidth]{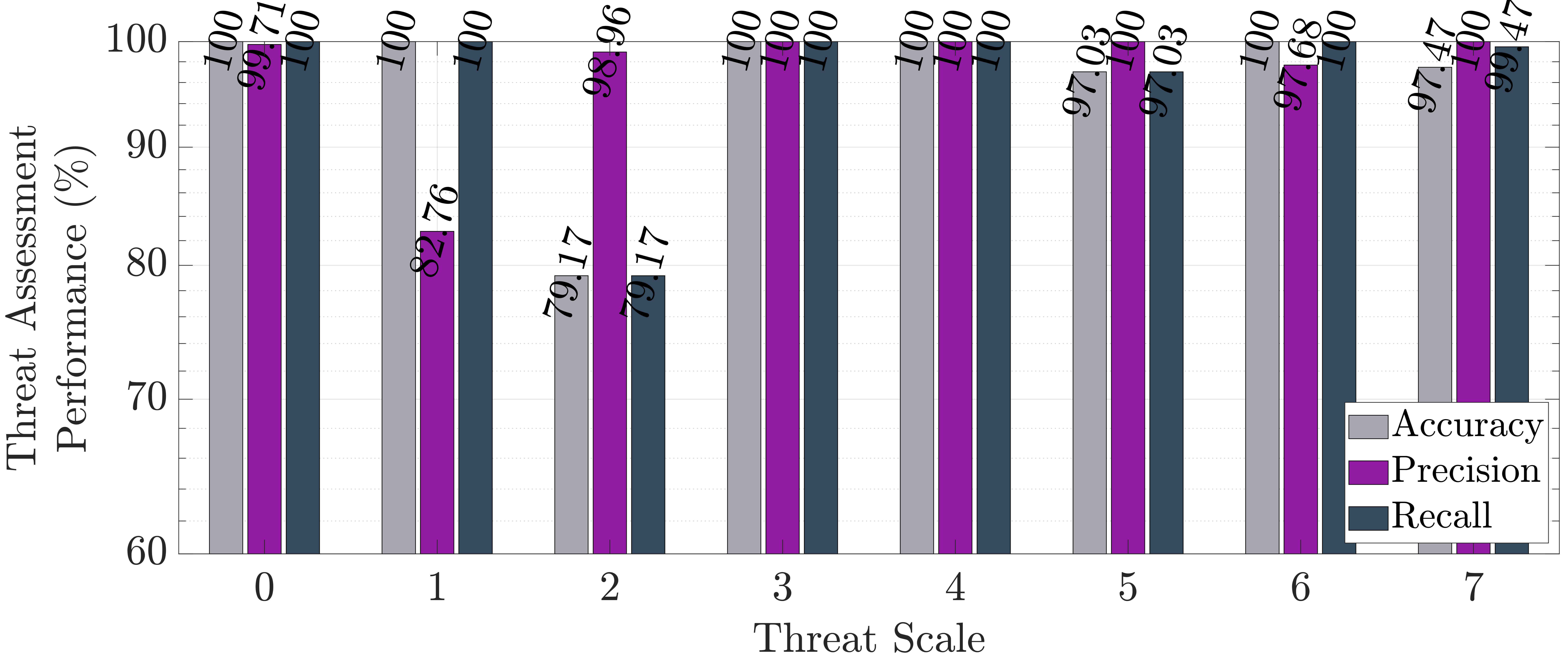}
\caption{Threat assessment results of the proposed CPA framework by scale.}
\label{fig4}
\end{figure}

Given the overall superiority of the multitask framework in Table~\ref{table4}, we now provide a more granular analysis of its threat assessment strategy for each threat scale, as illustrated in Fig. \ref{fig4}. The framework in most threat scales achieves superior performance with an accuracy of $100\%$ and proves its competence. However, it achieves an accuracy of $79.17\%$ that aligns with the recall at scale 2, while attaining accuracies and recalls of $97.03\%$ and $97.47\%$ for scales 5 and 7, respectively. These deviations due to false negatives reveal the challenges of assessing threats that have subtle differences in their characteristics in terms of capability and intent. Our framework is adept at addressing various threat behaviors while recognizing ambiguous circumstances that necessitate more analysis.

The performance of the multitask learning architecture and threat assessment strategy emphasizes the necessity of merging intent analysis with capability evaluation. Cooperating both tasks not only improves the detection accuracy for obvious cases, but also provides a robust mechanism to capture subtle discrepancies of threat models. By leveraging our feature representation algorithm, the CPA framework offers a comprehensive examination of threats that traditional methods might overlook.
\section{Conclusion}
We present a comprehensive CPA framework that leverages intent-driven threat assessment for enhanced space networks with intershell links. Through a multitask learning architecture that integrates capability and intent analyses of threats, our proposed framework overcomes security and reliability issues in communications. The results demonstrate that the framework achieves high accuracy across most threat scales, confirming its effectiveness in the threat detection and assessment. The benchmark study highlights how our CPA framework effectively overcomes the limitations of conventional sequential assessment methods. Overall, this letter provides a promising direction for improving the reliability and security of space communication systems, addressing vulnerabilities at cyber and physical levels.
\bibliographystyle{IEEEtran}
\bibliography{GecgelCetin_WCL2025-1438}
\end{document}